\pdfoutput=1

\documentclass[11pt]{article}

\usepackage[preprint]{acl}
\usepackage{subcaption}

\usepackage{times}
\usepackage{latexsym}
\usepackage{amsmath}
\usepackage{booktabs}
\usepackage{algorithm}
\usepackage{adjustbox}
\usepackage{algpseudocode}
\usepackage{xspace}
\usepackage{makecell}

\usepackage[T1]{fontenc}

\usepackage[utf8]{inputenc}

\usepackage{microtype}

\usepackage{inconsolata}

\usepackage{graphicx}

\usepackage{siunitx}
\title{
Large-Scale Data Selection for Instruction Tuning
}

\author{ \textbf{Hamish Ivison\textsuperscript{1,2}},
 \textbf{Muru Zhang\textsuperscript{1,3}},
 \textbf{Faeze Brahman\textsuperscript{2}},
\\
 \textbf{Pang Wei Koh\textsuperscript{1,2}}\textbf{,}
 \textbf{Pradeep Dasigi\textsuperscript{2}}
\\
 \textsuperscript{1}University of Washington,
 \textsuperscript{2}Allen Institute for AI,
 \textsuperscript{3}USC
\\
 \small{
   \href{mailto:hamishiv@cs.washington.edu}{hamishiv@cs.washington.edu}
 }}

\newcommand{\modelname}{\textsc{T\"ulu}\xspace}

\begin{document}
\maketitle
\begin{abstract}
Selecting high-quality training data from a larger pool is a crucial step when instruction-tuning language models, as carefully curated datasets often produce models that outperform those trained on much larger, noisier datasets.
Automated data selection approaches for instruction-tuning are typically tested by selecting small datasets ($\sim$10k samples) from small pools ($\sim$200k samples). However, popular deployed instruction-tuned models often train on hundreds of thousands to millions of samples, subsampled from even larger data pools. 
We present a systematic study of how well data selection methods scale to these settings, selecting up to 2.5M samples from pools of up to 5.8M samples and evaluating across 7 diverse tasks.
We show that many recently proposed methods fall short of random selection in this setting (while using more compute), and even decline in performance when given access to larger pools of data to select over.
We find that a variant of representation-based data selection (RDS+), which uses weighted mean pooling of pretrained LM hidden states, consistently outperforms more complex methods across all settings tested -- all whilst being more compute-efficient.
Our findings highlight that the scaling properties of proposed automated selection methods should be more closely examined.
\end{abstract}

\section{Introduction}



\begin{figure}
    \centering
    \includegraphics[width=\linewidth]{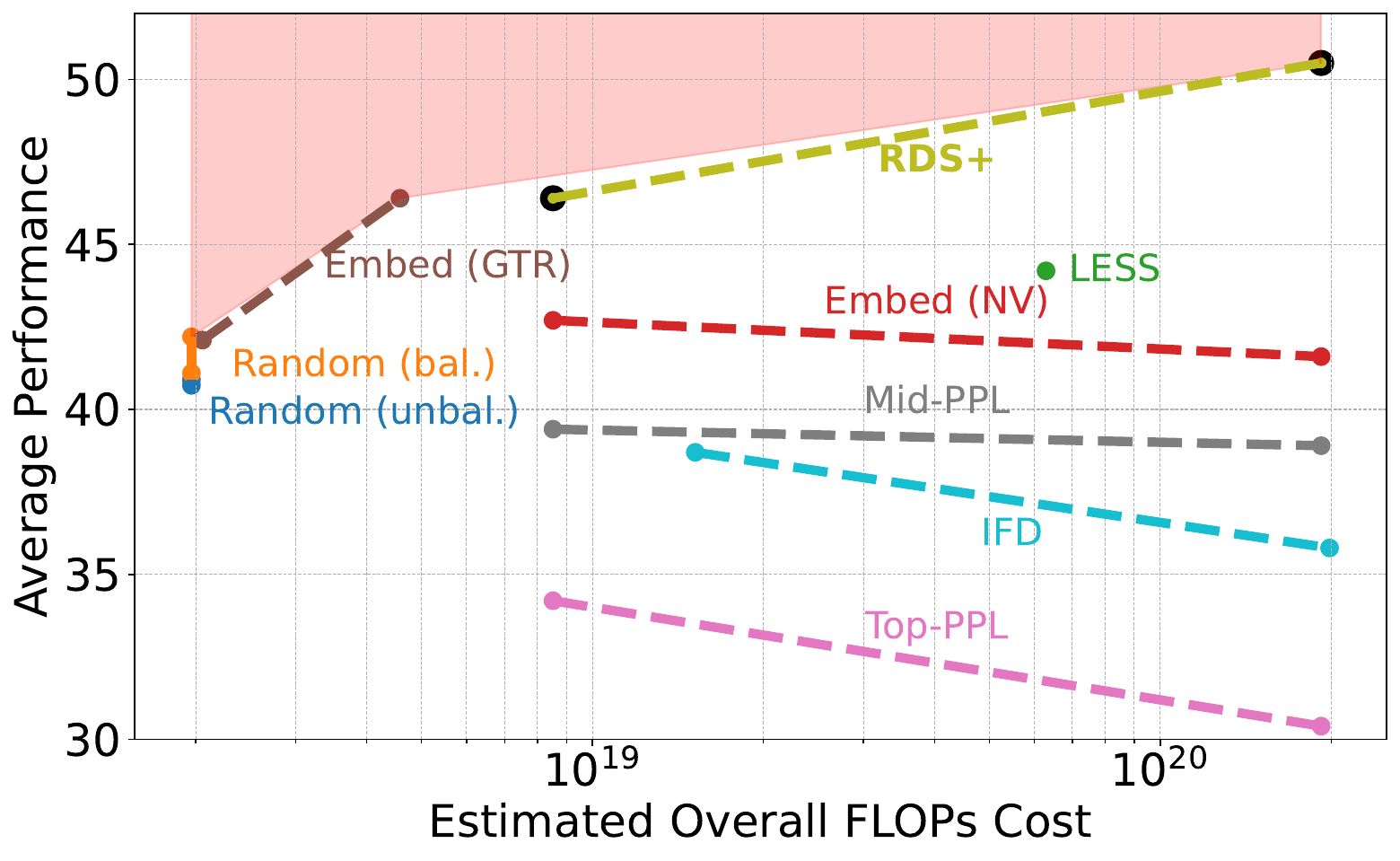}
    \caption{Performance against estimated compute cost of varied data selection methods when selecting 10k points from data pools consisting of 200k (left points) and 5.8M (right points) data points in the single-task setup described in \S\ref{sec:single_task_selection}.  We do not run LESS with 5.8M samples due to its high compute cost. Most data selection methods do not improve in performance with a larger pool, with the exception of RDS+ and Embed (GTR). We shade the Pareto frontier of efficiency and performance in red.}
    \label{fig:pareto}
\end{figure}

Instruction tuning has quickly become a crucial element of building modern language model (LM) systems, and a step for which curating high-quality data is especially important~\citep{zhou2023lima, wang2023how}. Finetuning on as few as 1,000 carefully chosen samples can yield models more preferred by human users than models trained on noisier datasets 10 times larger~\citep{zhou2023lima}.

The importance of data quality has led to a surge of papers proposing and investigating methods for selecting small (1–10k) datasets that outperform larger (50-200k) ones~\citep{chen2024alpagasus, xia2024less, li-etal-2024-quantity, liu2024what}. However, the sizes of the selected datasets are often significantly smaller than those of commonly used open post-training datasets, which typically contain hundreds of thousands to millions of examples\citep{ivison2023camelschangingclimateenhancing, lambert2024tulu3pushingfrontiers, grattafiori2024llama3herdmodels, deepseekai2025deepseekr1incentivizingreasoningcapability}.
As such, it is unclear how well these data selection approaches scale as both the number of the selected data points and size of the data pool grow.
Additionally, prior work suggests that common data selection approaches underperform random selection when considering the additional compute cost incurred by running the data selection techniques themselves even in smaller-scale settings~\citep{yin2024computeconstraineddataselection}.
It is unclear how performance and compute costs further change as we further scale beyond smaller-scale (selecting 10k data points from 200k samples) settings.
In this work, we aim to answer these questions by investigating how well data selection methods work for selecting instruction-tuning data when used to select larger datasets (100s of thousands of samples) from even larger data pools (millions of samples).

We construct data pools consisting of all data considered for \modelname~2~\citep{ivison2023camelschangingclimateenhancing} and \modelname~3~\citep{lambert2024tulu3pushingfrontiers}, state-of-the-art instruction-tuned models with publicly released instruction tuning data. 
These data pools are large, diverse, unbalanced, and of mixed quality, requiring automated data selection techniques that can scale to millions of samples, filter out noise, and maintain enough diversity to ensure effective generalization.
This allows us to examine how well data selection techniques scale up to selecting millions of samples from pools up to 6M data points in size.
We select for and evaluate on seven diverse evaluations, covering a diverse range of skills from code generation (HumanEval) to general chat (AlpacaEval). We additionally investigate selecting a single dataset for multiple tasks, reflecting how instruction-tuned models are often used to perform multiple tasks at once.

We test nine different data selection approaches in our setting, covering a variety of different approaches to data selection: gradient-based influence methods~\citep{xia2024less}, embedding-based methods \citep{zhang2018perceptual,hanawa2021evaluationsimilaritybasedexplanations,ivison-etal-2023-data}, and loss/perplexity-based methods~\citep{yin2024computeconstraineddataselection,antonello-etal-2021-selecting,marion2023moreinvestigatingdatapruning,ankner2024perplexedperplexityperplexitybaseddata, li-etal-2024-quantity}.
Surprisingly (and contrary to prior work), we find that a simple embedding method we call RDS+ --- using the hidden states of a pretrained LM --- works best overall across all settings tested, consistently beating all other approaches. In particular, we find that:

\begin{enumerate}
    \item \textbf{Many dataset selection methods struggle with increased dataset size.} 4 of the 7 methods we examine drop in performance as we increase the data pool they select over, even when selecting datasets of the same size. In contrast, RDS+ not only improves as the data pool size grows, but also \textbf{outperforms all other methods by at least 2 points on average}. This highlights that \textbf{examining the scaling properties of data selection is important for identifying strong methods} (\S\ref{sec:single_task_selection}).
    \item Similarly, when selecting for multiple tasks at once, \textbf{only 3 out of 7 selection methods beat random selection}, with RDS+ performing best. We find that RDS+ even outperforms \modelname~2, which was trained on a human-curated mixture (\S\ref{sec:multi_task_selection}). 
    \item Controlling for the compute used during selection, \textbf{RDS+ outperforms random selection by an average of two points} when selecting hundreds of thousands to millions of samples, whilst using less compute. This stands in contrast to prior work~\citep{xia2024less}, which does not scale up to large enough sizes to observe the point at which RDS becomes more compute-efficient. This highlights that \textbf{data selection techniques may only become particularly useful for larger-scale settings} than those typically used in prior work. (\S\ref{sec:scaling_multi_task_selection})
    
\end{enumerate}

We believe our findings highlight the importance of examining data selection methods at larger scales than considered in prior work, and show that some such methods can still provide benefits at these scales.


\section{Related Work}




\paragraph{Data Selection For Instruction Tuning} Earlier approaches to data selection for instruction tuning relied primarily on careful human curation~\citep{zhou2023lima} or extensive ablation studies to determine optimal data mixtures~\citep{wang2023how}. 
Recently, many automated data selection approaches have emerged, ranging from using GPT-based quality assessment~ \citep{chen2024alpagasus, wettig2024qurating, lu2023instag, liu2024what}, embedding or perplexity features from model inference~\citep{marion2023moreinvestigatingdatapruning, ivison-etal-2023-data,li-etal-2024-quantity, zhang2024automathtext, sachdeva2024traindataefficientllms, li-etal-2024-superfiltering, du2023mods, zhang-etal-2024-recost}, gradient-based methods \citep{xia2024less, han-etal-2020-explaining, zhang2024tagcostaskagnosticgradientclustered, yu2024mates}, or shallower heuristics~\citep{ZhaoACF24long, cao2024instructionmininginstructiondata, li-etal-2024-one}.
These works typically focus on finding small subsets of larger datasets that can match or outperform using the entire dataset, often selecting subsets of roughly 10k samples in size~\citep{chen2024alpagasus, xia2024less,li-etal-2024-quantity} from datasets with 100s of thousands of samples. This does not match the known sizes of various instruction-tuning datasets used in practice, which range from 300k to millions of samples in size~\citep{touvron2023llama2openfoundation, grattafiori2024llama3herdmodels, ivison2023camelschangingclimateenhancing, lambert2024tulu3pushingfrontiers}. In this work, we aim specifically to study how these methods perform when selecting large datasets from large pools, and find that many proposed methods fall short of random selection in this setting.

\paragraph{Surveying Data Selection Methods}

Some prior studies have similarly studied how well automated data selection techniques work for instruction tuning:
\citet{yin2024computeconstraineddataselection} examine how well various data selection approaches scale in compute-constrained scenarios. However, they use a relatively small data pool, only examine a single task setting, and assume only a single epoch of training (not common in instruction tuning).  \citet{diddee2024chasingrandominstructionselection} also examine how well data selection strategies generalize across different pools at various scales, but similarly only examine selecting relatively small datasets (up to 10k samples) and small pools (up to 200k in size). In contrast to both these works, we further scale the setting to use millions of samples, more evaluations, and more diverse datasets, and find that this changes our view of what method works best, with RDS outperforming random selection even in FLOPs-controlled scenarios.

\citet{dai2024improving} examine how to balance influence scores when selecting multi-task data, and uses a selection algorithm similar to our proposed round-robin approach. However, we also examine how well multi-task selection works when selecting from significantly larger data pools, across multiple different pools and base models.

\section{Data Selection Methods}
\label{sec:data_selection_methods}

We now describe a set of popular data selection methods commonly used and tested in prior work alongside the experimental setup we use to investigate them. We assume that one has a pool of data $D$ and wishes to select up to $n$ instances from the pool. 
Additionally, we assume that the user has some (10s to 100s) of query samples from a query set $V$. This query set is a small dataset that is in the same distribution as the evaluation set (e.g., we can use existing validation sets as query sets if available). We assume both $V$ and $D$ contain prompts and responses (i.e., both are labelled). In practice, this query split can either be task-specific (i.e., some query split of the downstream task we want to test on), or contain samples from a variety of tasks that do not overlap with downstream test tasks, but are reflective of our desired test distribution in some way.
Each method either (a) takes in a dataset $D$ as input and produces a score for each point $d \in D$ or (b) takes in a dataset $D$ and query points $v \in V$ and produces a score for each $v, d \in V, D$ pair, which we then aggregate to compute a score for each $d \in D$ (described further in \S\ref{sec:select_aggregate}).
Given this, we explore the following methods. We pick a set of methods that represent simple but varied ways to select data: using model embeddings, using loss-based features (perplexity, IFD), using model gradients, and random selection. Our choices match common baselines used in prior work~\citep{xia2024less,yin2024computeconstraineddataselection}.

\paragraph{Random Selection.} We explore random selection, a common strong baseline for data selection. We report both taking random samples of a given data pool, and taking `balanced' random samples, where we uniformly sample from data sources. If we run out of data from a given source, we divide its remaining `budget' equally among the dataset sources we have not yet exhausted.

\paragraph{Length.} We sort examples by length (in tokens) and take the longest samples. This has been shown to be a strong baseline by \citet{longismore}, and is computationally cheap, only requiring computing the length of each sample in our data pool.

\paragraph{Perplexity.} 
We compute the loss of each $d \in D$ on our original base model and use it as its score,  following prior work~\citep{yin2024computeconstraineddataselection,antonello-etal-2021-selecting,marion2023moreinvestigatingdatapruning,ankner2024perplexedperplexityperplexitybaseddata}. We use the same setup as \citet{yin2024computeconstraineddataselection} and examine both `mid-ppl' (taking points in the middle of the score distribution) and `top-ppl' (taking only the highest loss points).

\paragraph{IFD.} We follow the procedure used in \citet{li-etal-2024-quantity}, which involves first training a model on representative samples from the dataset, and then scoring data points using the ratio of the answer loss given the question to the loss of the answer on its own (called the IFD score). We compute the IFD score for all points $d \in D$. We use the codebase provided by the authors.\footnote{\url{https://github.com/tianyi-lab/Cherry_LLM}} When selecting data from the 5.8M \modelname 2 unfiltered pool, we use the same model trained on the 200k-size pool, as the smaller pool is simply a subsampled version of the larger one.

\paragraph{LESS.} We follow the procedure outlined in \citet{xia2024less}, training LoRAs on a random subset of the data, and then selecting data by computing the gradient-based influence of each $d \in D$ on each $v \in V$ to obtain the selection score for the pair $v, d$. We use the codebase provided by \citet{xia2024less}.\footnote{\url{https://github.com/princeton-nlp/LESS}}

\paragraph{Embedding.} We use an embedding model to score each pair $v, d \in V, D$ based on the cosine similarity of $v$ and $d$ when processed by the embedding model. We test using two different embedding models: NV-Embed-v2~\citep{lee2024nvembedimprovedtechniquestraining}, which was at the top of the MTEB leaderboard~\citep{muennighoff-etal-2023-mteb} at time of experimentation, and GTR-base~\citep{ni-etal-2022-large}, following \citet{yin2024computeconstraineddataselection}.

\paragraph{RDS+.} Finally, similar to embedding-based models, we explore using the hidden representations of the model we plan to train, as opposed to a trained embedding  model~\citep{zhang2018perceptual,hanawa2021evaluationsimilaritybasedexplanations,ivison-etal-2023-data, xia2024less} -- representation-based data similarity (RDS).
We use a custom variant of RDS in which we take a position-weighted mean pool of the last hidden layer states (see \S\ref{sec:ablations} for details).
We compute this for each $v, d$ in $V$ and $D$ and compute the cosine similarity for each pair $v, d \in V, D$.
We ablate alternate variants of RDS in App.~\ref{sec:ablations}, and denote our tuned version of RDS as \textbf{`RDS+'}.

\subsection{Selection \& Aggregation}
\label{sec:select_aggregate}

\begin{algorithm}[t]
\caption{Single-task selection method
}\label{alg:round_robin}
\begin{algorithmic}
\Require A dictionary of scores $S$ such that $S[v,d]$ is the score given by a selection method between points $v \in V$ and $d \in D$.
\Require $n$, the desired size of our selected dataset.
\State $L \gets []$
\While{$|L| < n$}
\For{$v \in V$}
\State $L \gets \text{argmax}_{d \in D}{S[v,d]}$
\State $S[v,d] \gets -\inf$ \Comment{Set score such that this point will not get chosen again.}
\If{$|L| \geq n$}
break
\EndIf
\EndFor
\EndWhile
\State \Return $L$
\end{algorithmic}
\end{algorithm}

For methods that provide a score for each pair $v, d \in V, D$, we end up with $|V|$ scores for each data point $d \in D$. As such, we need to determine how to aggregate these scores to determine the most valuable points $d \in D$. In pilot experiments, we found using a round-robin approach worked best, iteratively adding to the selected pool the highest-scoring point for each $v \in V$ until we reach a desired size. This is what we use to construct task-specific datasets.
We illustrate the algorithm in Alg~\ref{alg:round_robin}.

For multi-task scenarios, we also need to aggregate on a task-level when combining scores from different tasks. We say that $V_t$ is the query set for task $t$, and $S[v_t, d]$ is the score computed between dataset point $d$ and query point $v_t$.
We then construct a dictionary of scores $S'$ where $S[t,d]$ represents the score of a $d$ for task $t$ by setting $S[t,d] = \max_{v_t \in V_t}{S[v_t, d]}$.
We then run a round-robin procedure where we iterative over tasks, `pop' out the highest-scoring data point, add to our dataset, and repeat until we have a dataset of the desired size (after deduplication).
This is essentially running Alg.~\ref{alg:round_robin} again, but replacing $S$ with $S'$ and $V$ with $T$.
We also experimented with averaging task-level scores together, but found this performed worse overall (See App.~\ref{app:sel_algo_multitask} for details).


\subsection{Data Pool}

\begin{figure*}
    \centering
    \includegraphics[width=.9\linewidth]{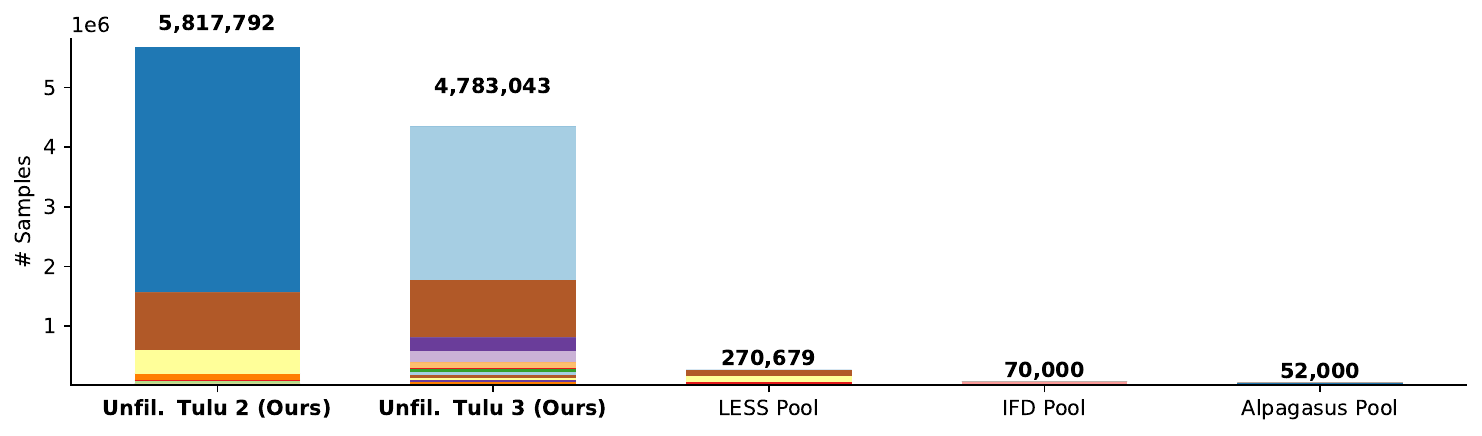}
    \caption{Size and makeup of data pools considered in this work (unfiltered Tulu 2, 3) and in past work~\citep{xia2024less,chen2024alpagasus,li-etal-2024-quantity}. We provide the size of each pool on top of each bar. Each color represents a different dataset. See App.~\ref{app:tulu_3_unfiltered} for more details on data pool composition and exact per-source counts.}
    \label{fig:datapool_mixture}
\end{figure*}

In this work, we wish specifically to explore how well automated data selection methods work \textbf{at scale}. As such, we test the above methods in three different settings, across two different dataset pools: \modelname~2 unfiltered and \modelname~3 unfiltered. Both pools comprise of millions of samples with diverse data sources, and are constructed by examining the original datasets chosen for \modelname 2 and 3 (state-of-the-art post-trained models with openly available instruction data) and retrieving the same datasets but performing no downsampling. We perform exact-match deduplication of samples over the pool to ensure all samples are unique. Note that the unfiltered mix is extremely unbalanced, comprising mostly of data from FLAN and Open Orca. We compare the makeup and size of our data pools to pools in prior work~\citep{xia2024less, li-etal-2024-quantity} in Fig.~\ref{fig:datapool_mixture}.
Notably, our pools are significantly larger and more diverse than prior work, and are based on real datasets used for open instruction-tuned models.

\subsection{Experimental Design}

We extend our experimental design off \modelname~2~\citep{ivison2023camelschangingclimateenhancing}, an open-source state-of-the-art dataset and model family (at time of release). As \modelname~2 is finetuned starting from Llama 2 base models~\citep{touvron2023llama2openfoundation}, we primarily experiment with the Llama 2 7B model.
We additionally report results using the \modelname 3 mixture~\citep{lambert2024tulu3pushingfrontiers}, used to train a state-of-the-art open instruction-tuned model, and Llama 3.1~\citep{grattafiori2024llama3herdmodels} in \S\ref{sec:selecting_from_other_data_pools}, and 8 varied models in \S\ref{sec:analysis}.


\paragraph{Training.} For finetuning, we fully finetune for two epochs with a batch size of 1, 128 gradient accumulation steps, a learning rate of \num{2e-5} (\num{1e-5} for 70B size models), 
linear learning rate warmup for the first 3\% of training steps, and linear cooldown for the rest of training. This follows the settings used in prior work~\citep{wang2023how,ivison2023camelschangingclimateenhancing}. We report the mean across three random runs (including reselecting data) for random baselines and single-run scores for other settings.

\paragraph{Evaluation.} For evaluation, we largely follow the same evaluation procedures as described in \citet{wang2023how, ivison2023camelschangingclimateenhancing}. We evaluate on (and select for) MMLU~\citep{hendrycks2020measuring}, GSM8k~\citep{cobbe2021gsm8k}, BBH~\citep{suzgun2022challenging}, TydiQA~\citep{clark-etal-2020-tydi}, HumanEval-Codex~\citep{chen2021codex}, Squad~\citep{rajpurkar-etal-2016-squad}, and AlpacaEval~\citep{alpaca_eval}. We either use the existing few-shot examples or create custom development splits from the evaluations for data selection. We also experiment with using out-of-domain query sets in \S\ref{sec:multi_task_selection}.
We provide precise details on each evaluation and query set in Appendix~\ref{app:eval_details}.

\paragraph{FLOPs Estimation.} For the FLOPs estimates used throughout the paper, we follow \citet{kaplan2020scalinglawsneurallanguage} in estimating the compute cost of a training step as roughly $6N$ FLOPs per token processed and an inference step as $2N$ per token processed, where $N$ is the parameter count of the model. Based on this, we estimate the FLOPs of each method using the total number of inference and training steps that take place across selection and model training. We provide further details in App.~\ref{app:flops_details}.

\section{Results}

\subsection{Single-Task Data Selection}
\label{sec:single_task_selection}

\begin{table*}[t]
\centering
\adjustbox{max width=.95\linewidth}{
\begin{tabular}{lcccccccc}
\toprule
\textbf{Method} & \textbf{MMLU} & \textbf{GSM8k} & \textbf{BBH} & \textbf{TydiQA} & \textbf{Codex} & \textbf{Squad} & \textbf{AlpacaEval} & \textbf{Average} \\
\midrule \multicolumn{9}{c}{Selecting 10k datapoints for single tasks from 200k data points} \\ \midrule
Random (unbal.)    & 45.8 & 17.2 & 41.5 & 52.7 & 27.0 & 83.5 & 18.4 & 40.9 \\
Random (bal.)      & 45.4 & 16.6 & 40.7 & 50.7 & 27.9 & 80.6 & 33.2 & 42.2 \\
LESS               & 46.7 & \textbf{21.5} & 41.8 & \textbf{57.5} & 19.6 & 84.7 & 37.9 & 44.2 \\
Embed (NV)         & 46.9 & 19.9 & 42.4 & 52.3 & 25.0 & \textbf{88.3} & 24.0 & 42.7 \\
Embed (GTR)        & \textbf{47.1} & 20.2 & 42.4 & 52.4 & 27.7 & \textbf{88.3} & 16.7 & 42.1 \\
Top-PPL            & 46.4 &  9.9 & 33.8 & 46.9 & 20.3 & 77.6 & 4.3 & 34.2 \\
Mid-PPL            & 46.9 & 15.5 & 39.7 & 53.6 & 25.7 & 81.6 & 12.9 & 39.4 \\
IFD                & 42.6 & 17.5 & 39.1 & 38.7 & 29.1 & 57.9 & 45.9 & 38.7 \\
Length & 43.2 & 14.2 & 40.2 & 54.3 & 30.4 & 82.3 & 40.0 & 43.5 \\
\textbf{RDS+}              & 47.0 & 20.9 & \textbf{42.7} & 52.3 & \textbf{31.8} & 88.2 & \textbf{42.3} & \textbf{46.4} \\ \midrule
\multicolumn{9}{c}{Selecting 10k datapoints for single tasks from all 5.8M data points} \\ \midrule
Random (unbal.) & 46.9 & 17.4 & 42.1 & 53.2 & 26.8 & 83.6 & 15.1 & 40.7 \\
Random (bal.) & 45.3 & 16.5 & 40.2 & 50.9 & 28.2 & 78.0 & 29.0 & 41.1 \\
Top-PPL            & 44.6 & 6.1 & 23.9 & 40.4 & 20.9 & 74.4 & 2.5 & 30.4 \\
Mid-PPL            & 46.1& 14.9 & 40.7 & 52.7 & 23.0 & 82.3 & 12.5 & 38.9 \\
Embed (GTR) & 45.0 & 29.9 & 42.8 & 45.5 & 29.7 & 88.3 & 43.8 & 46.4 \\
Embed (NV) & 47.0 & 23.1 & 41.1 & 51.2 & 29.1 & \textbf{88.8} & 10.9 & 41.6 \\
IFD & 41.9 & 13.0 & 37.9 & 40.0 & 26.4 & 46.6 & 44.6 & 35.8 \\
Length & 40.7 & 2.4 & 39.2 & 17.8 & 20.9 & 0.2 & 3.1 & 17.8 \\
\textbf{RDS+} & \textbf{47.5} & \textbf{33.9} & \textbf{42.9} & \textbf{54.9} & \textbf{32.4} & 88.5 & \textbf{53.5} & \textbf{50.5} \\
\bottomrule
\end{tabular}}
\caption{Single-task performance of different data selection techniques over the \modelname~2 unfiltered set. Each cell reports the performance of a model trained with 10k samples chosen for that particular target task.
We show results selecting from a downsampled form or full set of the \modelname~2 unfiltered set.
We find RDS performs best overall, even beating more computationally expensive methods like LESS.}
\label{tab:single_task_results_10k}
\end{table*}


We start by testing the different data selection methods over \modelname 2 unfiltered both with a smaller pool (\modelname~2 unfiltered downsampled to 200k samples) and the entire data pool. We select 10k samples, train models, and then evaluate for each task separately. We report the results in Table~\ref{tab:single_task_results_10k}.
We do not report LESS performance using the 5.8M pool as it requires computing gradients three times over all 5.8M samples in the pool, which is beyond our computational budget.

\paragraph{RDS+ performs best.} RDS+ is the highest-performing method on average in both settings, despite being similar or cheaper in cost than most other baselines (Fig.~\ref{fig:pareto}). Additionally, models trained with RDS-selected data perform the best in each task individually (with the exception of SQuAD, where they are a close second) when selecting from the full pool.

\paragraph{PPL, Random, IFD, and Embed (NV) perform worse with a larger pool.} As seen in Fig.~\ref{fig:pareto}, multiple methods we test actually perform \textit{worse} when we select over the larger data pool, suggesting that they cannot effectively scale. In contrast, both RDS+ and Embed (GTR) improve with a larger pool. In later areas of the paper, we focus on RDS+ as we wish to achieve the strongest possible performance with a larger compute budget.



\subsection{Multi-task Selection}
\label{sec:multi_task_selection}

\begin{table*}[ht]
\centering
\adjustbox{max width=\linewidth}{
\begin{tabular}{lcccccccc}
\toprule
\textbf{Method} & \textbf{MMLU} & \textbf{GSM8k} & \textbf{BBH} & \textbf{TydiQA} & \textbf{Codex} & \textbf{Squad} & \textbf{AlpacaEval} & \textbf{Average} \\
\midrule
Rand. (unbal) & \textbf{52.2} & 18.0 & 44.5 & 55.3 & 25.7 & 81.5 & 33.9 & 44.5 \\
Rand. (bal) & 51.5 & 21.8 & \textbf{45.1} & 50.7 & 32.2 & 87.9 & 43.2 & 47.5 \\
Top-PPL            & 49.1 & 10.5 & 39.4 & 49.4 & 21.6 & 80.3 & 5.6 & 36.6 \\
Mid-PPL            & 53.1& 13.3 & 42.8 & 52.4 & 20.3 & 86.2 & 20.7 & 41.3 \\
Embed (GTR) & 49.9 & 32.8 & 44.6 & 54.4 & 30.4& 88.4 & 35.7 & 48.0 \\
Embed (NV) & 50.6 & 28.7 & 44.4 & 56.0 & 30.4 & 89.1 & 17.9 & 45.3 \\
IFD & 45.7 & 21.8 & 41.2 & 39.5 & 27.7 & 17.0 & 57.4 & 35.7 \\
Length & 50.0 & 16.4 & 38.8 & 54.9 & 25.0 & \textbf{89.3} & 58.5 & 47.6 \\
\modelname 2 & 50.0 & 22.7 & \textbf{45.1} & 54.0 & 33.1 & 86.9 & 54.4 & 49.5 \\
\textbf{RDS+}  & 50.2 & 35.2 & 44.7 & 56.3 & \textbf{35.1} & 89.0 & 45.6 & \textbf{50.9} \\ \midrule
\textbf{RDS+}  - Wildchat & 50.9 & 24.8 & 43.6 & \textbf{57.3} & 31.1 & 87.3 & 39.3 & 47.8 \\
\textbf{RDS+}  - Arena Hard & 48.1 & \textbf{36.2} & 43.9 & 51.8 & 31.8 & 81.3 & \textbf{59.4} & 50.4 \\
\bottomrule
\end{tabular}}
\caption{Multi-task performance of dataset selection methods when selecting 326k samples from the full \modelname 2 unfiltered pool. 
Each row reflects the performance of a single model trained on a single dataset chosen to perform well across tasks.
For `WildChat' and `Arena Hard' we use samples from WildChat and Arena Hard for selection.
}
\label{tab:multi_task_results_320k}
\end{table*}

We next examine how well data selection methods work when selecting \textit{single} datasets while targeting \textit{multiple} tasks. We use the round-robin method described in \S\ref{sec:select_aggregate} to balance selection across tasks for task-specific methods\footnote{We found this worked best across multiple data selection methods.}, or just use the existing scores for task-agnostic methods. We select 326k samples from the \modelname~2 unfiltered set, matching the number of samples in the \modelname~2 SFT mixture, and so also compare to \modelname~2.
We show the results in Table~\ref{tab:multi_task_results_320k} and find that:

\paragraph{RDS+ still performs best overall.} RDS+ outperforms all other methods on average, including \modelname~2 itself, showing that \textbf{RDS+ can human-curated mixtures}. In fact, we observe that all methods except RDS+ and Embed (GTR) still underperform random selection, suggesting that embedding-based methods overall are best for data selection.

\paragraph{RDS+ selection performs well even when the evaluations are out of distribution.} We explore using WildChat~\citep{zhao2024wildchat} and Arena Hard~\citep{li2024crowdsourced} as out of distribution  query sets for selecting data.
Crucially, this means \textit{we do not assume access to any query samples from our test tasks}. We find that using Arena Hard samples performs close to RDS+, showing we can select high-quality samples without assuming any data from the evaluations in our test suite. This suggests that using a high-quality selection set results in a data mixture that generalizes well to unseen tasks.

\paragraph{RDS+ performs strongly with other data pools.}
\label{sec:selecting_from_other_data_pools}
In order to test how well our findings generalize, we make use of the recently released state-of-the-art \modelname~3 data and model \citep{lambert2024tulu3pushingfrontiers}, comparing the model trained with RDS-selected data the officially released \modelname~3 SFT model. We then select 939k instances, equal to the size of \modelname~3 SFT Mix, from the pool and finetune Llama 3.1 models on it using the hyperparameters from \citet{lambert2024tulu3pushingfrontiers}. We show our results in Table~\ref{tab:llama_3_results}. We find that \textbf{RDS+ round-robin selected data outperforms the official \modelname~3 SFT checkpoint} and random baselines, showing that RDS+ remains an effective data selection method even on a different base model and with a different selection data pool.

\subsection{Scaling Multi-task Selection}
\label{sec:scaling_multi_task_selection}
\begin{table*}[t]
\centering
\adjustbox{max width=\linewidth}{
\begin{tabular}{lcccccccc}
\toprule
\textbf{Method} & \textbf{MMLU} & \textbf{GSM8k} & \textbf{BBH} & \textbf{TydiQA} & \textbf{Codex} & \textbf{Squad} & \textbf{AlpacaEval} & \textbf{Average} \\
\midrule
Random (unbal.) & 61.6 & \textbf{81.2} & 66.8 & 71.1 & 76.4 & 89.7 & 75.6 & 74.6 \\
Random (bal.) & 62.1 & 76.0 & 68.6 & 68.8 & \textbf{87.2} & 87.4 & 72.4 & 74.7 \\
\modelname~3 SFT & 62.2 & 74.3 & \textbf{68.2} & 67.4 & 83.8 & 85.5 & 71.9 & 73.3 \\ \midrule
\textbf{RDS+}  & \textbf{62.5} & 77.6 & 66.6 & \textbf{72.1} & 83.8 & \textbf{90.2} & 80.2 & \textbf{76.1} \\
\textbf{RDS+}  - Arena Hard & 57.0 & 78.7 & 59.7 & 49.4 & 75.7 & 66.3 & \textbf{84.5} & 67.3 \\
\bottomrule
\end{tabular}}
\caption{Multi-task performance of RDS against baselines when finetuning from Llama 3.1 8B base and selecting 939k samples from the \modelname 3 unfiltered mixture following the multitask setup in \S\ref{sec:multi_task_selection}.
For `Arena Hard' we use samples from Arena Hard as the query set.
RDS+ outperforms the official \modelname 3 SFT model.}
\label{tab:llama_3_results}
\end{table*}

\begin{figure}[t]
    \centering
    \includegraphics[width=\linewidth]{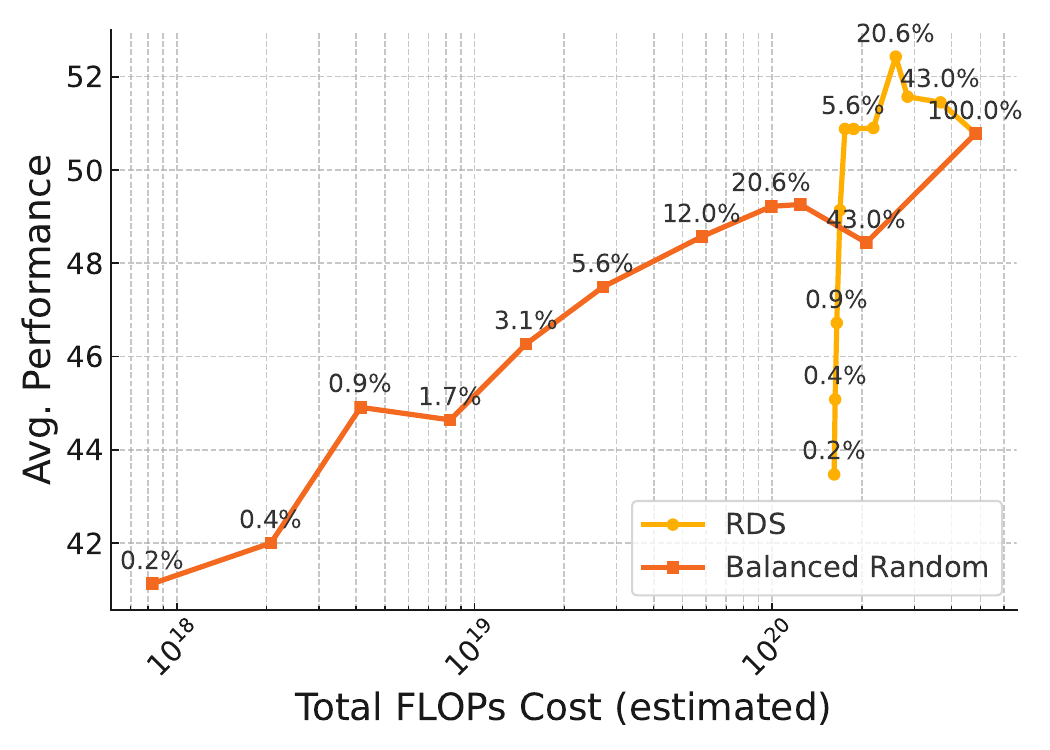}
    \caption{Average multi-task performance against FLOPs cost (including selection) for balanced random and RDS+. We label points with the \% of the total data pool used. RDS+ outperforms random selection significantly when selecting less data, and is more FLOPs efficient at larger selection sizes. See App.~\ref{app:flops_details} for details on FLOPs estimates.}
    \label{fig:flops_against_performance}
\end{figure}

\begin{figure}
    \centering
    \includegraphics[width=\linewidth]{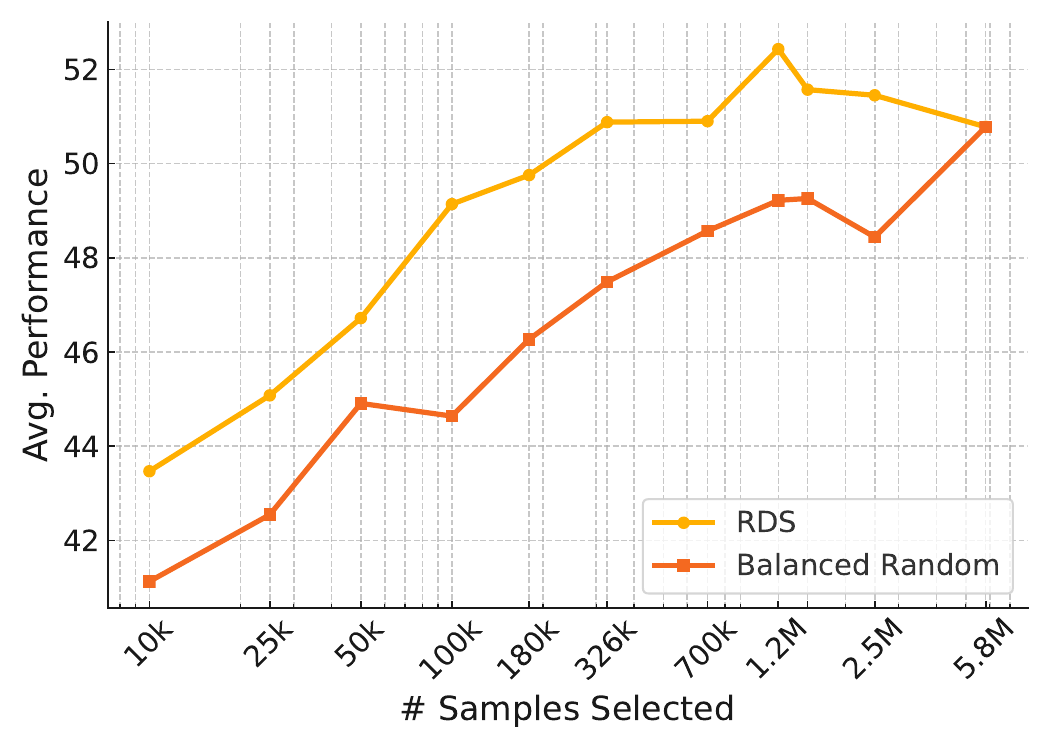}
    \caption{Average multi-task performance against number of samples selected. RDS+ consistently beats balanced random at all data sizes tested, up to using the entire data pool.}
    \label{fig:rds_data_scaling}
\end{figure}

Finally, we examine how the performance of our selection method changes as we scale the amount of data being selected.
Due to the high cost of these experiments (fully finetuning on millions of samples), we only look at RDS+ (the best-performing method overall) and random selection (a strong, cheap baseline).
We examine RDS+ as the strongest performing data selection method, and compare to random selection as a cheap but often effective baseline (that also outperforms all but two of the tested data selection approaches).
We select for multiple tasks at once from the full \modelname 2 unfiltered pool, selecting varying numbers of samples with RDS+.
We plot average performance against the number of samples selected in Figure~\ref{fig:rds_data_scaling}. We also plot performance against total FLOPs used for selection and training in Figure~\ref{fig:flops_against_performance}. We observe that:

\paragraph{RDS+ consistently beats balanced random selection across different data selection sizes} (Fig.~\ref{fig:rds_data_scaling}). This shows that targeted selection is even useful when selecting datasets up to millions of data points in size. Furthermore, RDS+ performs similarly to training on all data (50.8 vs 50.9) even when selecting only 326k samples (roughly 6\% of the overall pool), and \textbf{outperforms training on all data when selecting over 1M samples}. 

\paragraph{RDS+ can achieve better performance with less compute when selecting more data points} (Fig.~\ref{fig:flops_against_performance}).
When taking into account the compute used for selection and training, we find that the extra compute incurred by RDS does not pay off until we select relatively large datasets (roughly $\geq 326$k), after which it becomes significantly better than random selection.
This highlights that examining data selection performance at scale is important to show the potential of some methods.


\section{Analysis}
\label{sec:analysis}

\paragraph{What gets selected?}
We examine qualitatively what samples get selected by RDS+, IFD, and PPL (see App.~\ref{app:selected_viz} for a visualization).
For RDS+, We find that the sources selected vary depending on downstream evaluation, often with intuitive explanations: e.g., GSM8k requires strong chain-of-thought reasoning abilities, and so the chain-of-thought data is upweighted relative to the normal distribution. Similarly, for HumanEval-Codex the Code Alpaca dataset is upweighted.

Additionally, we observe that PPL and IFD appear to select more ShareGPT and FLAN data respectively than RDS+ or random selection, suggesting these methods have biases to particular types of data. This also explains  why IFD performs relatively well in AlpacaEval (which improves when training on GPT-derived data), but drops in other aspects, while Top-PPL achieves very low AlpacaEval scores in Table~\ref{tab:single_task_results_10k}.


\paragraph{Can we reduce the cost of RDS+?}
While we have used the same model for selection and training for RDS+, we also wished to investigate to what extent changing the model used for selection impacts data selection performance, potentially allowing RDS+ to use less or more compute for selection.
Encouragingly, we find that using smaller models for selection (even from different model families) can still results in strong performance, suggesting that RDS+ can indeed be made computationally cheaper by using smaller models for selection.
However, we do not observe gains from using larger models compared to Llama 2 7B, suggesting that using larger models for selection does not yield improved performance.
We provide further details in App.~\ref{app:varying_model_size_family}.


\section{Conclusion}

In this work, we have explored how well a variety of data selection techniques work when selecting datasets of up to 2.5M samples from pools comprising of close to 6M data points.
By examining data selection techniques in these settings, we find that many methods proposed in the past not only underperform random selection baselines, but also perform worse with larger pools of data.
Embedding-based methods, and in particular RDS+, are the primary exception, outperforming all other data selection methods across various selection sizes.
Finally, we find that using smaller models to select data can also work well, suggesting future work could further reduce the cost of RDS+ by exploring how best to use smaller proxy models.
We believe that our results overall highlight the importance of testing data selection methods over large, diverse pools of data, and testing how well data selection approaches scale with respect to both data and compute. We will publicly release our data and code to aid in future work.



\section*{Limitations}

While we cover a reasonable number of base models in this work, we ultimately only use two data pools, \modelname~2 and \modelname~3, due to the computational cost of running experiments over millions of instances. We hope to further examine how well our findings transfer to data pools with different characteristics in future work.

Additionally, we note that any data selection method that requires model passes over data will scale in cost proportionally with the data pool used to select, which limits the use of this method in extremely large-scale settings -- where even doing model forward passes over all data may be too computationally expensive. While this is less common in instruction-tuning settings (as there is less instruction-tuning data available generally), it may happen for pretraining datasets containing trillions of tokens (which is not a focus in this work).

Finally, while we do not explicitly analyze the safety and potential toxicity of our models, we hope that our findings could be used to improve data selection for selecting safety-relevant data, which often does make up some portion of popular instruction data mixes~\citep{lambert2024tulu3pushingfrontiers}. Furthermore, we hope that reducing the total FLOPs cost of training strong instruction models can help aid the overall environmental cost of training LMs~\citep{energy_nlp}.

\bibliography{iclr2025_conference,anthology}

\appendix
\section{Evaluation Details}
\label{app:eval_details}
We provide more details on each evaluation setting and how we constructed the split used for data selection below. In all cases we use an instruction chat template following the \modelname~format, matching the format used during SFT training.

\begin{enumerate}
    \item \textbf{MMLU}~\citep{hendrycks2020measuring}: We use the official MMLU evaluation script and prompts available at \url{https://github.com/hendrycks/test}. We evaluate using 0 few-shot examples, following the original setup of MMLU. We report average accuracy across test examples. For the query set, we use the dev examples (commonly used as few-shot samples for evaluation).
    \item \textbf{GSM8k}~\citep{cobbe2021gsm8k}: We evaluate models on the test set of GSM. Following \citet{wei2022chain}, we evaluate with chain-of-thought. We use 8 few-shot in-context examples. Because all answers in GSM are numbers, we extract the last number in the model response as the final answer. We report average accuracy across test examples. For the query set, we use the 8 few-shot examples individually (without the other shots included in the prompt).
    \item \textbf{Big Bench Hard (BBH)}~\citep{srivastava2023beyond, suzgun2022challenging}:  We follow the setup described in the original paper and evaluate with chain-of-thought. The officially provided prompts, which have 3 few-shot in-context examples are used. For the CoT setup, we extract the first word after the phrase `So the answer is', or the entire response if there is no such substring present. We report average accuracy over sub-tasks (all of which use accuracy as the primary metric). For the query set, we use the 3 few-shot examples from each task individually (without the other shots included in the prompt).
    \item \textbf{TydiQA}~\citep{clark-etal-2020-tydi}: We follow the setup described in the PaLM 2 technical report~\citep{anil2023palm} to evaluate models' performance in answering multilingual questions under two settings when the gold passage that contains the answer is given (i.e., gold passage setting). One in-context example is used to familiarize the model with the answering format. For the query set, we use the 9 few-shot examples from each language individually (without any shots included in the prompt).
    \item \textbf{HumanEval Codex}~\citep{chen2021evaluating}: We use the HumanEval dataset in the codex paper for evaluating models' coding ability. The dataset contains 164 programming problems, where models are prompted to complete the Python function given its docstring. Following the original paper, we compute unbiased estimates of pass@k to measure the functional correctness of models' outputs. We report pass@10, sampling with a temperature of 0.8. We additionally use the instructions provided by HumanEvalPack~\citep{muennighoff2023octopack}, as this better suites instruction-tuned models. We create a custom test split of 148 samples, and evaluate on those. We use the remaining 16 samples as the query split.
    \item \textbf{SQuAD}~\citep{rajpurkar-etal-2016-squad}: We use the validation split of SQuAD as a test set, comprising of 10,570 questions about Wikipedia articles. We include the article containing the answer in the prompt, and include 3 in-context examples (randomly selected from the train set) in order to ensure the model outputs in the desired format.\footnote{In pilot experiments, we found that models without in-context samples would often provide the right answer in a verbose manner, harming the string-matching-based metrics used for SQuAD.} We report text-based F1. We use 500 random samples from the SQuAD train set as query samples.
    \item \textbf{AlpacaEval}~\citep{alpaca_eval}: We use the package provided by \citet{alpaca_eval}, following the default setup for both AlpacaEval 1. We allow the evaluated model to generate up to 8192 tokens, without specifying special stop sequences. We create a custom split of 50 samples for the query set, and use the remaining samples for evaluation.
\end{enumerate}

\begin{table}[t]
    \centering
    \begin{tabular}{lcc}
        \toprule
        \textbf{Dataset} & \textbf{\# Query} & \textbf{\# Test} \\
        \midrule
        \textbf{MMLU}      & 285   & 14,042 \\
        \textbf{GSM8k}     & 8     & 1,319 \\
        \textbf{BBH}       & 81    & 6,511 \\
        \textbf{TydiQA}    & 9     & 5,077 \\
        \textbf{Codex}     & 16    & 148 \\
        \textbf{Squad}     & 500   & 10,570 \\
        \textbf{AlpacaEval} & 50   & 755 \\
        \bottomrule
    \end{tabular}
    \caption{Query and test split counts for evaluation datasets.}
    \label{tab:eval_counts}
\end{table}

We provide a summary of the number of query and test samples in Table~\ref{tab:eval_counts}.

\section{Data Pool Sources Breakdown}
\label{app:tulu_3_unfiltered}

\begin{table*}
\centering
\setlength\tabcolsep{5pt}
\adjustbox{max width=\linewidth}{
\begin{tabular}{@{}lccccc@{}}
\toprule
\textbf{Source} & \textbf{\modelname~2} & \textbf{\modelname~2 Unfil.} & \textbf{LESS} & \textbf{IFD} & \textbf{Alpagasus} \\ \midrule
FLAN V2~\citep{flant5} & 49,123 & 961,322 & 100,000 & - & - \\
FLAN CoT~\citep{flant5} & 49,747 & 398,439 & 100,000 & - & - \\
Open Assist.~\citep{kopf2023openassistant} & 7,331 & 7,707 & 55,668 & - & - \\
Dolly~\citep{dolly} & 0 & 15,007 & 15,011 & - & - \\
GPT-4 Alpaca~\citep{peng2023instruction} & 19,906 & 52,002 & - & - & - \\
Code Alpaca~\citep{codealpaca} & 20,016 & 20,022 & - & - & - \\
ShareGPT & 111,912 & 100,054 & - & - & - \\
LIMA~\citep{zhou2023lima} & 1,018 & 1,030 & - & - & - \\
Wizard Evol-Instruct V2~\citep{xu2023wizardlm} & 29,810 & 142,802 & - & - & - \\
Open Orca~\citep{OpenOrca} & 29,683 & 4,111,858 & - & - & - \\
SciRIFF~\citep{sciriff} & 7,468 & 7,535 & - & - & - \\
Alpaca~\citep{alpaca} & - & - & - & 52,002 & 52,002 \\
WizardLM (70K)~\citep{xu2023wizardlm} & - & - & - & 70,000 & - \\
Hardcoded & 140 & 14 & - & - & - \\ \midrule
\textbf{Total} & 326,153 & 5,817,792 & 270,679 & 122,002 & 52,002 \\ \bottomrule
\end{tabular}
}
    \caption{Number of samples per dataset in the original \modelname~2 dataset and our unfiltered version alongside the data pools used in LESS~\citep{xia2024less}, IFD~\citep{li-etal-2024-quantity}, and Alpagasus~\citep{chen2024alpagasus}. Note that we deduplicate samples in the \modelname~2 unfiltered mixture.}
    \label{tab:filtered_set_details}
\end{table*}

\begin{table*}[t]
\centering
\begin{tabular}{lcc}
\toprule
\textbf{Source} & \textbf{\# Samples in \modelname~3} & \textbf{\# Samples in Unfil.} \\
\midrule
\modelname~3 Hardcoded & 240 & 24 \\
Open Assist.~\citep{kopf2023openassistant} & 7,132 & 7,131 \\
No Robots~\citep{no_robots} & 9,500 & 9,500 \\
WildChat (GPT-4 subset)~\citep{zhao2024wildchat} & 100,000 & 235,028 \\
FLAN V2~\citep{flant5} & 89,982 & 961,322 \\
SciRIFF~\citep{sciriff} & 10,000 & 35,149 \\
TableGPT~\citep{zha2023tablegptunifyingtablesnature} & 5,000 & 13,159 \\
\modelname~3 Persona MATH & 149,960 & 149,960 \\
\modelname~3 Persona GSM & 49,980 & 49,980 \\
\modelname~3 Persona Algebra & 20,000 & 50,000 \\
OpenMathInstruct 2~\citep{toshniwal2024openmath2} & 50,000 & 2,570,505 \\
NuminaMath-TIR~\citep{numina_math_datasets} & 64,312 & 64,312 \\
\modelname~3 Persona Python & 34,999 & 34,998 \\
Evol CodeAlpaca~\citep{luo2024wizardcoder} & 107,276 & 107,276 \\
\modelname~3 CoCoNot~\citep{brahman-kumar2024} & 10,983 & 10,983 \\
\modelname~3 WildJailbreak~\citep{jiang2024wildteaming} & 50,000 & 178,344 \\
\modelname~3 WildGuardMix~\citep{wildguard2024} & 50,000 & 85,090 \\
Aya~\citep{singh-etal-2024-aya} & 100,000 & 190,320 \\
\modelname~3 Persona IF & 29,980 & 29,962 \\
\midrule
\textbf{Total} & 939,344 & 4,783,043 \\
\bottomrule
\end{tabular}
\caption{
Number of samples per dataset in the original
\modelname~3 SFT mixture and our unfiltered version.
See \citet{lambert2024tulu3pushingfrontiers} for further details on all splits.}
\label{tab:tulu_3_splits}
\end{table*}

We provide a breakdown of the splits of the \modelname~2 SFT mixture and our `unfiltered' version, along with other data pools from prior work, in Table~\ref{tab:filtered_set_details}. We provide a similar comparison between the \modelname~3 SFT mixture and our `unfiltered' version in Table~\ref{tab:tulu_3_splits}.

\section{Compute Details}

We run all experiments on a cluster containing H100s and A100s with up to 8 GPUs per node. We find that training on 10,000 examples takes roughly 30 minutes on 1 H100 GPU, while training on 2.5 million samples (our largest selected set) takes 62 hours on a node of 8 H100s. Running the selection itself varies according to method, with RDS itself taking 87 H100 GPU-hours for indexing the \modelname~3 unfiltered set (although this can be efficiently parallelized).

\section{Multi-task Selection Algorithm}
\label{app:sel_algo_multitask}

\begin{table}[t]
\centering
\adjustbox{max width=\linewidth}{
\begin{tabular}{llc}
\toprule
\textbf{Sel. Meth.} & \textbf{Aggr. Meth.} & \textbf{Avg. Perf.} \\
\midrule
RDS+ & round-robin & \textbf{50.9} \\
RDS+ & mean-max & 47.9 \\ \midrule
Embed (GTR) & round-robin & \textbf{48.0} \\
Embed (GTR) & mean-max & 44.9 \\
\bottomrule
\end{tabular}}
\caption{Average performance of different methods when selecting 326k samples from \modelname 2 unfiltered using the multitask setting, using either round-robin or mean-max methods for aggregating samples across tasks. Round-robin beats mean-max performance in both cases. See App.~\ref{app:sel_algo_multitask} for details.}
\label{tab:meanmax_vs_rr}
\end{table}

\begin{figure}[t]
    \centering
    \includegraphics[width=.5\textwidth]{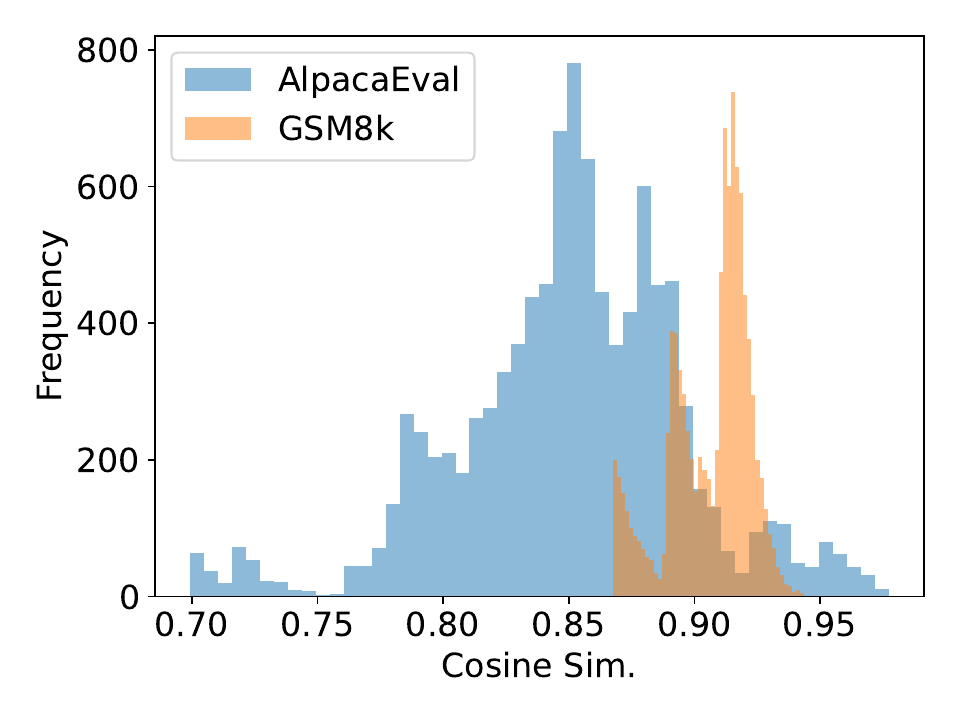}
    \caption{Histogram of RDS scores for the top 10,000 samples picked for GSM8k and AlpacaEval from the \modelname~2 unfiltered pool. We find that AlpacaEval instances have lower average similarity than GSM8k.}
    \label{fig:scores_dist}
\end{figure}

We experimented with two methods for selecting multi-task datasets: `round-robin' (the method explained in \S~\ref{sec:select_aggregate}) and `mean-max'. For `mean-max', we compute the per-task scores for $d$ in the same way as before, but then simply average all task scores for a datapoint together. Using the notation from \S~\ref{sec:select_aggregate}, we do: $S[d] = \frac{\sum_{t \in T} S[t, d]}{|T|}$, where $S[d]$ is the score for $d$. We then can simply take the top-scoring points as our dataset.

We compare using round-robin and mean-max for RDS+ and Embed in Table~\ref{tab:meanmax_vs_rr}. We find that round-robin consistently outperforms mean-max across methods on average, and so we use it in our core experiments.

We further investigate why this is the case by visualizing the distribution of RDS scores over the \modelname~2 unfiltered pool in Figure~\ref{fig:scores_dist} for GSM8k and AlpacaEval.
We observe that scores for GSM8k are consistently higher than AlpacaEval, suggesting that GSM8k scores `dominate' when averaging task scores together.
This means that the final selected dataset is likely to be mostly comprised of samples that are useful for GSM8k, but not AlpacaEval. We find similar trends comparing AlpacaEval to other datasets, suggesting that merely averaging tasks scores leads to some targeted evaluations being under-served.
This agrees with concurrent work suggesting that round-robin algorithms that carefully balance selected samples across tasks outperform more naive methods~\citep{dai2024improving}.

\section{Further Details on FLOPs Estimates}
\label{app:flops_details}

For the FLOPs estimates used throughout the paper, we follow \citet{kaplan2020scalinglawsneurallanguage} in estimating the compute cost of a training step as roughly $6N$ FLOPs per token processed, where $N$ is the parameter count of the model (roughly 7B). \citet{kaplan2020scalinglawsneurallanguage} notes that the forward pass is roughly half the cost of the backward pass, giving us an estimate of $2N$ FLOPs per token when processing samples. We use a rough estimate of 2,048 tokens per sample, since during training and selection we truncate all samples to be at most this length. Note we fully-finetune models for two epochs in all setups. Let $N$ be the model size, $P$ be the size of the data pool (in number of samples), and $D$ the number of samples selected to train on. Based on this, the cost for each method is estimated as follows:
\begin{enumerate}
    \item \textbf{Random Selection}: $2 * 2048 * 6ND$
    \item \textbf{Perplexity}: $2 * 2048 * 2NP + 2 * 2048 * 6ND$
    \item \textbf{IFD}: $200000 * 2049 * 2N + 1000 * 2048 * 6ND + 2 * 2048 * 2NP + 2 * 2048 * 6ND$ (We train the initial model used to compute IFD scores on 1000 samples selected from the 200k data pool.)
    \item \textbf{LESS}: $3 * 2048 * 6NP + 2 * 2048 * 6ND$ (LESS computes gradients for three checkpoints over the entire pool.)
    \item \textbf{Embedding}: $2 * 2048 * 2NP + 2 * 2048 * 6ND$
    \item \textbf{RDS+}: $2 * 2048 * 2NP + 2 * 2048 * 6ND$
\end{enumerate}

Note that for embedding and RDS+, we can select using a smaller model than we train (as done for Embed (GTR) or the experiments in App.~\ref{app:varying_model_size_family}). In this case, we adjust the inference cost computation accordingly.
We assume the cost of processing scores is negligible compared to the rest of the procedure, since in practice this runs on CPU in under an hour for methods we test.
These formulations provide some intuition for why testing large selection sets (large $P$) is important: if $P >> D$, methods like RDS+ use significantly more compute than random selection. As $D$ approaches $P$, the added cost of doing inference over the data pool becomes less significant.



\section{Varying Model Size and Family for RDS+}
\label{app:varying_model_size_family}

\begin{table*}[t]
\centering
\begin{adjustbox}{width=\textwidth}
\begin{tabular}{lcccccccc|c}
\toprule
\makecell[c]{\textbf{Train Model $\rightarrow$} \\ \textbf{Sel. Model $\downarrow$}} & \textbf{Q2.5} & \textbf{L2 7B} & \textbf{L2 13B} & \textbf{L2 70B} & \textbf{L3.1 8B} & \textbf{L3.1 70B} & \textbf{O2 8B} & \textbf{O2 13B} & \textbf{Avg.} \\
\midrule
\textbf{Qwen 2.5 1.5B} & \textbf{56.5} & 43.5 & 54.4 & 70.6 & 65.8 & 82.7 & 57.0 & 64.6 & 61.9 \\
\textbf{Llama 2 7B}     & 52.0 & 45.1 & 57.0  & 72.4 & \textbf{70.4} & 83.1 & 57.5 & \textbf{66.7} & \textbf{63.0} \\
\textbf{Llama 2 13B}    & 52.0 & 42.4 & 54.4  & 70.3 & 62.9 & 82.3 & 56.3 & 63.4 & 60.5 \\
\textbf{Llama 2 70B}    & 53.7 & \textbf{45.4} & 54.3  & 72.3 & 68.4 & \textbf{83.2} & 57.1 & 64.9 & 62.4 \\
\textbf{Llama 3.1 8B}   & 53.0 & 45.0 & 55.9  & \textbf{72.5} & 67.8 & 82.1 & 57.2 & 65.7 & 62.4 \\
\textbf{Llama 3.1 70B}  & 53.2 & 45.1 & \textbf{57.3}  & 71.7 & 68.0 & 82.1 & \textbf{57.6} & 65.3 & 62.5 \\
\textbf{Olmo 2 7B}      & 53.2 & 43.5 & 54.6  & 71.4 & 66.9 & 82.2 & 57.4 & 64.8 & 61.8 \\
\textbf{Olmo 2 13B}     & 51.0 & 43.2 & 54.9  & 71.0 & 65.8 & 83.0 & 57.1 & 65.1 & 61.4 \\
\bottomrule
\end{tabular}
\end{adjustbox}
\caption{
Average multi-task performance of RDS round-robin when when varying the model doing the selecting and the model being trained. We find that using a different model to the one being trained does not hurt performance, with Llama 2 7B being the best selector overall.}
\label{tab:sel_train_vary}
\end{table*}

Since RDS+ relies on the hidden states of the model, we examine to what extent changing the model used for selection impacts data selection performance. We construct a pool of 100k samples from \modelname~2 unfiltered, balanced via uniform random downsampling across the different sources. We select from the pool using RDS+ with a range of models varying both in size (from 1.5B to 70B) and in family (Qwen~\citep{qwen2.5}, Llama 2~\citep{touvron2023llama2openfoundation}, Llama 3.1~\citep{grattafiori2024llama3herdmodels}, OLMo 2~\citep{olmo2}), and show the results in Table~\ref{tab:sel_train_vary}. Surprisingly, we find that using selection models from different families to the ones being trained can still result in strong performance, suggesting that RDS' strong performance does not come from matching the selection and downstream model, but from good general selection of samples.
While we do observe some outliers (Llama 2 7B and 13B being particularly good and bad respectively), this suggests that using larger selection models or matching the selection and training model is not crucial for RDS+. This promising finding means that we can potentially select data for much larger models with much smaller ones, significantly reducing the compute cost of RDS+.





\section{Further Details on RDS+}

\subsection{Ablations}
\label{sec:ablations}

\begin{table}[t]
\centering
\begin{tabular}{lc}
\toprule
\textbf{Method} & \textbf{Avg. Perf.} \\
\midrule
RDS (\textbf{ours}) & \textbf{46.4} \\ \midrule
- EOS token only & 45.4 \\
- Uniform mean-pool & 45.8 \\ \midrule
- Prompt-only & 43.2 \\
- Label-only & 45.1 \\
\bottomrule
\end{tabular}
\caption{Overall performance of RDS variations when selecting 10k samples for single target tasks (matching Tab.~\ref{tab:single_task_results_10k}).
Using only labels is surprisingly effective.}
\label{tab:rds_ablations}
\end{table}

In this work, we use a custom RDS variant (`RDS+') that we chose based on a series of ablations testing (a) using different sets of hidden states and (b) using different parts of the input data itself. We perform the ablations in the 200k single-task setting (matching the setting used for Table~\ref{sec:single_task_selection}).

For constructing the embeddings, we considered using just the hidden state corresponding to the final EOS token (EOS token only)~\citep{xia2024less}, mean-pooling the final hidden states across the sequence (uniform mean pool), or using a weighted mean-pooling approach that takes into account the causal attention mask used with decoder-only models, inspired by past work on converting decoder-only models to generic text embedding models~\citep{muennighoff2022sgptgptsentenceembeddings}. See below for further detail on the weighting. As shown in Table~\ref{tab:rds_ablations}, using the EOS token alone or uniform mean-pooling underperforms our chosen weighted mean pooling approach.
Second, we explore using just the initial user turn (prompt-only) or the only the first response (label-only) instead of the entire input. Surprisingly, using only the label outperforms using only the prompt, although both underperform using the entire sample, despite the label often containing relatively little task information (e.g., just the letter answer for multiple-choice questions).

\subsection{RDS+ Weighted Pooling Details}
\label{app:weighting}

For the weighted mean-pooling strategy, we follow \citet{muennighoff2022sgptgptsentenceembeddings} in using position weighting to average hidden states across the model inputs while taking into account the causal mask. Specifically, token $i$ has weight $w_i$ as follows:
\[w_i = \dfrac{i}{\sum_{i=1}^{L} i}\]
Where $L$ is the total length in tokens of the given sample. Given these weights, we then simply perform weighted averaging to compute the RDS embedding:
\[\text{embedding} = \sum_{i=1}^{L}w_ih_i\]
Where $h_i$ is the last layer hidden state of the $i^\text{th}$ token.
The rationale for this weighting is due to the causal mask of the decoder-only models we use for embedding: since the model is trained with a causal mask, the first input token does not see the rest of the sequence. Likewise, the second token does not see token 3 onwards, etc. As such, naive mean-pooling may bias the features towards earlier tokens in the sequence, since hidden states only accumulate features from previous tokens. The weighted mean-pooling strategy attempts to counter this by then weighting later tokens heavier.

\section{Visualization of RDS+}

\begin{figure*}
    \centering
    \includegraphics[width=\linewidth]{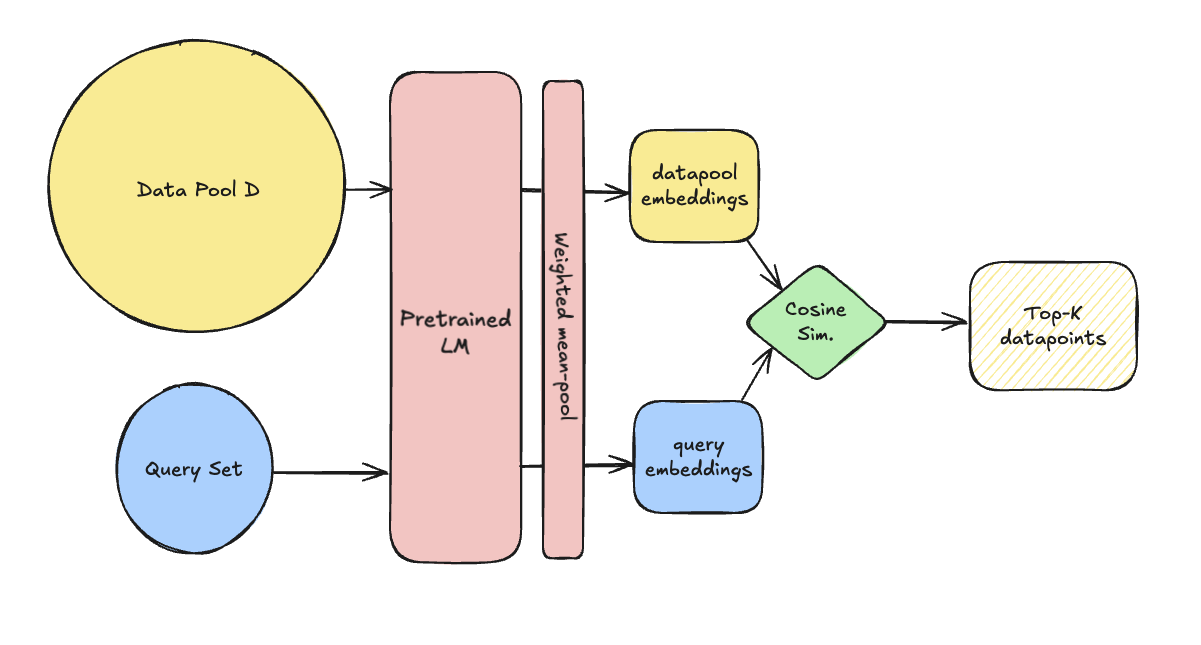}
    \caption{Overview of RDS+ data selection. A pretrained LM encodes a data pool and a set of query points via weighted mean-pooling. We then compute cosine similarity between each point in the pool and the query set, and pick the top-K most similar points using Algorithm~\ref{alg:round_robin}.}
    \label{fig:rds_overview_app}
\end{figure*}

We provide a basic visualization of RDS+ in Figure~\ref{fig:rds_overview_app}.

\section{Visualization of Selected Samples}
\label{app:selected_viz}
\begin{figure*}[t]
    \centering
    \includegraphics[width=\linewidth]{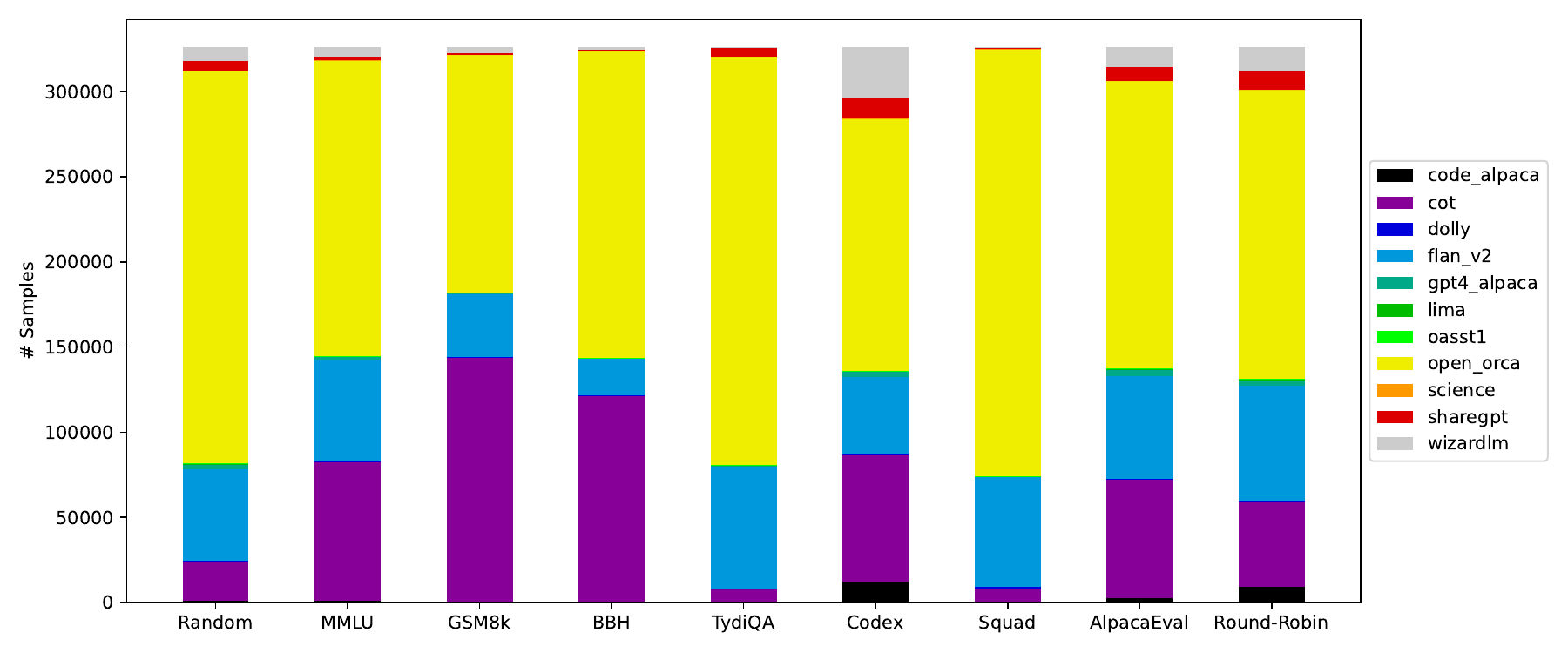}
    \caption{Breakdown of what data gets selected when selecting 326,000 samples using RDS from the \modelname~2 unfiltered pool. `Random' represents the samples chosen when randomly downsampling to 326,000 samples, and `round-robin' refers to the samples selected by the multi-task round-robin selection.}
    \label{fig:selected_viz}
\end{figure*}

\begin{figure*}[t]
    \centering
    \includegraphics[width=\linewidth]{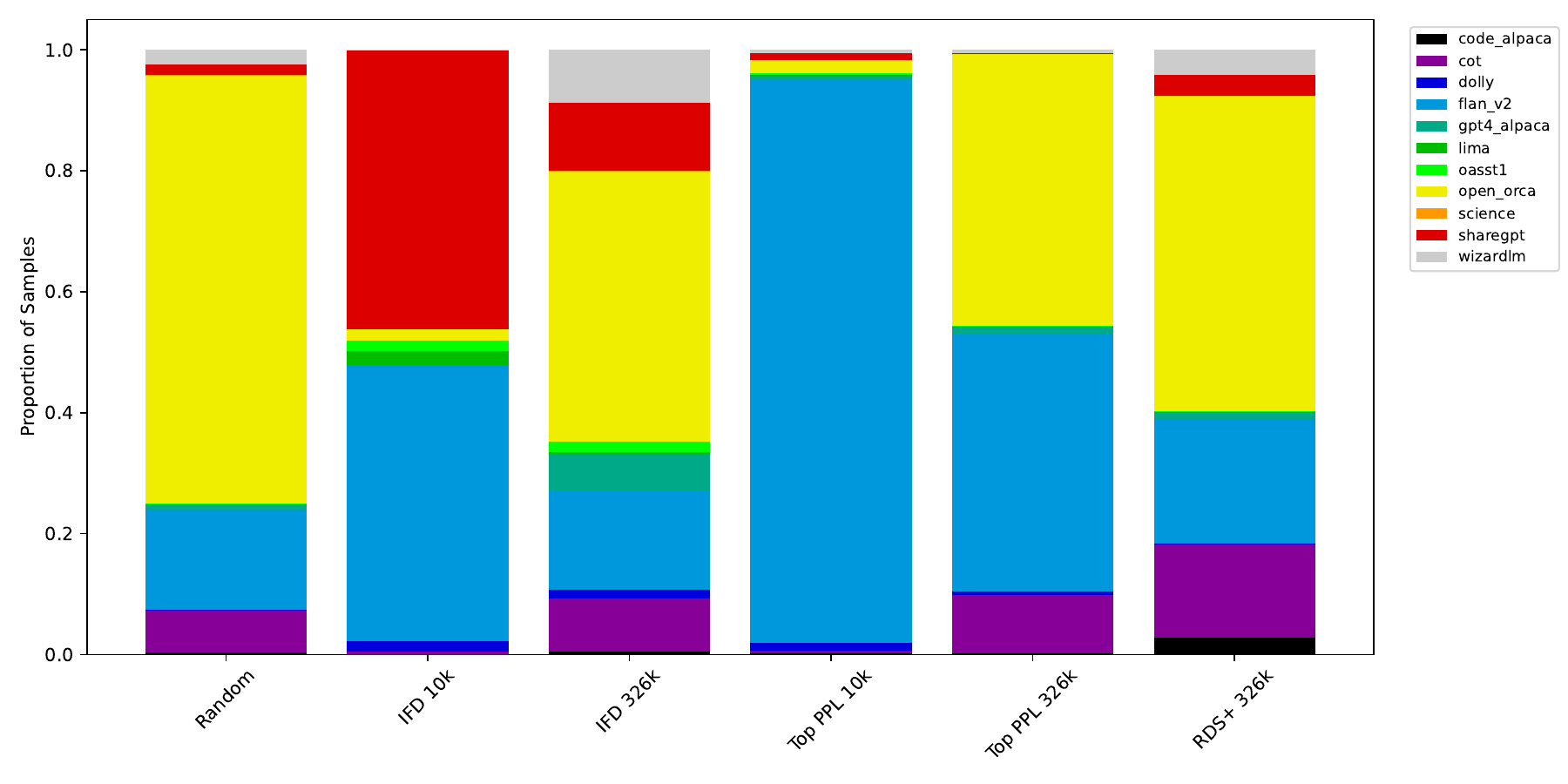}
    \caption{Breakdown of what data gets selected when selecting 10,000 or 326,000 samples using RDS from the \modelname~2 unfiltered pool using various selection methods. Sample counts normalized to add to 1. `Random' represents the samples chosen when randomly downsampling to 326,000 samples. IFD has a clear bias to ShareGPT data at both sizes, while PPL has a clear bias to FLAN data.}
    \label{fig:selected_viz2}
\end{figure*}

We visualize what samples get selected when selecting 326,000 samples using RDS+ from the \modelname~2 unfiltered pool in Figure~\ref{fig:selected_viz}, and what samples get selected when selecting with IFD, Top-PPL, random (unbalanced), and RDS+ from the \modelname~2 unfiltered pool in Figure~\ref{fig:selected_viz2}. We provide the raw per-source counts used to construct these figures in Tables~\ref{tab:selected_viz} and \ref{tab:dataset_viz2_table} respectively.

\begin{table*}[h]
\centering
\small
\adjustbox{width=\linewidth}{\begin{tabular}{lrrrrrrrrr}
\toprule
\textbf{Dataset} & \textbf{AlpacaEval} & \textbf{BBH} & \textbf{Codex} & \textbf{GSM8k} & \textbf{MMLU} & \textbf{SQuAD} & \textbf{TyDiQA} & \textbf{Round-robin} & \textbf{Random} \\
\midrule
\texttt{code\_alpaca} & 2386 & 38 & 12231 & 28 & 963 & 32 & 0 & 9084 & 1126 \\
\texttt{cot}          & 69859 & 121838 & 74385 & 144253 & 81532 & 8349 & 7482 & 50481 & 22396 \\
\texttt{dolly}        & 272 & 35 & 63 & 68 & 103 & 818 & 0 & 288 & 862 \\
\texttt{flan\_v2}     & 60381 & 21098 & 45531 & 36379 & 59818 & 64291 & 72724 & 67275 & 53851 \\
\texttt{gpt4\_alpaca} & 4189 & 677 & 3609 & 1391 & 2377 & 562 & 48 & 3389 & 2864 \\
\texttt{hard\_coded}  & 0 & 0 & 0 & 0 & 0 & 0 & 0 & 0 & 1 \\
\texttt{lima}         & 46 & 6 & 50 & 14 & 26 & 49 & 1 & 59 & 56 \\
\texttt{oasst1}       & 222 & 12 & 199 & 22 & 48 & 59 & 416 & 545 & 439 \\
\texttt{open\_orca}   & 168766 & 180159 & 147830 & 139379 & 173252 & 250736 & 239301 & 170031 & 230329 \\
\texttt{science}      & 0 & 1 & 528 & 0 & 173 & 95 & 0 & 9 & 440 \\
\texttt{sharegpt}     & 8300 & 440 & 12248 & 869 & 2324 & 656 & 5879 & 11140 & 5568 \\
\texttt{wizardlm}     & 11579 & 1696 & 29326 & 3597 & 5384 & 353 & 149 & 13699 & 8068 \\
\bottomrule
\end{tabular}}
\caption{Raw dataset counts used to create Figure~\ref{fig:selected_viz}, showing the breakdown of what data gets selected when selecting 326,000 samples using RDS from the \modelname~2 unfiltered pool. }
\label{tab:selected_viz}
\end{table*}
\begin{table*}[t]
\centering
\begin{tabular}{lrrrrrr}
\toprule
\textbf{Dataset} & \textbf{Random} & \textbf{IFD 10k} & \textbf{IFD 326k} & \textbf{Top PPL 10k} & \textbf{Top PPL 326k} & \textbf{RDS+ 326k} \\
\midrule
\texttt{flan\_v2}       & 961322 & 4559  & 53693  & 9305  & 138898 & 67275 \\
\texttt{cot}            & 398439 & 50     & 29025  & 46     & 31289  & 50481 \\
\texttt{oasst1}         & 7707   & 177    & 5476   & 27     & 293     & 545 \\
\texttt{dolly}          & 15007  & 172    & 4382   & 127    & 1751   & 288 \\
\texttt{gpt4\_alpaca}   & 52002  & 1      & 19914  & 96     & 3964   & 3389 \\
\texttt{code\_alpaca}   & 20022  & 3      & 1362   & 15     & 854     & 9084 \\
\texttt{sharegpt}       & 100054 & 4606  & 36582  & 109    & 597     & 11140 \\
\texttt{lima}           & 1030   & 234    & 861     & 1      & 32      & 59 \\
\texttt{wizardlm}       & 142802 & 11     & 28618  & 58     & 1699   & 13699 \\
\texttt{open\_orca}     & 4111858 & 187    & 146064 & 216    & 146622 & 170031 \\
\texttt{science}        & 3684   & 0      & 23      & 0      & 1       & 9 \\
\bottomrule
\end{tabular}
\caption{Raw dataset counts used to create Figure~\ref{fig:selected_viz2}, showing the breakdown of what data gets selected when selecting 10,000 or 326,000 samples using RDS from the \modelname~2 unfiltered pool using various selection methods.}
\label{tab:dataset_viz2_table}
\end{table*}


\end{document}